\documentclass[preprint,times,review]{elsarticle}

\usepackage{graphicx}
\usepackage{amsmath}
\usepackage{amssymb}
\usepackage{booktabs}

\usepackage{multirow}
\usepackage{bm}
\usepackage{color}

\newenvironment{packed_itemize}{
	\begin{itemize}
		\setlength{\itemsep}{1pt}
		\setlength{\parskip}{0pt}
		\setlength{\parsep}{0pt}
	}{\end{itemize}}

\usepackage{xspace}
\makeatletter
\DeclareRobustCommand\onedot{\futurelet\@let@token\@onedot}
\def\@onedot{\ifx\@let@token.\else.\null\fi\xspace}

\def\eg{\emph{e.g}\onedot} 
\def\ie{\emph{i.e}\onedot}

\def\etal{\emph{et al}\onedot}
\makeatother

\usepackage{amsmath}
\DeclareMathOperator*{\argmax}{arg\,max}

\usepackage{booktabs,subcaption,amsfonts,dcolumn}
\newcolumntype{d}[1]{D..{#1}}

\usepackage[pagebackref,breaklinks,colorlinks]{hyperref}

\usepackage[capitalize]{cleveref}
\crefname{section}{Sec.}{Secs.}
\Crefname{section}{Section}{Sections}
\Crefname{table}{Table}{Tables}
\crefname{table}{Tab.}{Tabs.}

\journal{Pattern Recognition}

\begin{document}
	\captionsetup[figure]{name={Fig.}}
	\begin{frontmatter}

		\title{Classification of Single-View Object Point Clouds}

        \author[scut]{Zelin Xu}
        \ead{eexuzelin@mail.scut.edu.cn}
        \author[scut]{Ke Chen\corref{cor1}}%
        \ead{chenk@scut.edu.cn} 
        \author[scut]{Kangjun Liu}
        \ead{wikangj.liu@mail.scut.edu.cn}
        \author[scut]{Changxing Ding}
        \ead{chxding@scut.edu.cn}
        \author[pcl]{Yaowei Wang}
        \ead{wangyw@pcl.ac.cn}
        \author[scut]{Kui Jia\corref{cor1}}%
        \ead{kuijia@scut.edu.cn}
		
		\cortext[cor1]{Corresponding author.}
        \address[scut]{South China University of Technology, Guangzhou 510641, China}
        \address[pcl]{Peng Cheng Laboratory, Shenzhen 518000, China }
		
		\begin{abstract}
        Object point cloud classification has drawn great research attention since the release of benchmarking datasets, such as the ModelNet and the ShapeNet.
        These benchmarks assume point clouds covering \emph{complete} surfaces of object instances, for which plenty of high-performing methods have been developed. However, their settings deviate from those often met in practice, where, due to (self-)occlusion, a point cloud covering \emph{partial} surface of an object is captured from an arbitrary view.
        We show in this paper that performance of existing point cloud classifiers drops drastically under the considered \emph{single-view, partial} setting; the phenomenon is consistent with the observation that semantic category of a partial object surface is less ambiguous only when its distribution on the whole surface is clearly specified. 
        To this end, we argue for a single-view, partial setting where supervised learning of object pose estimation should be accompanied with classification. Technically, we propose a baseline method of \emph{Pose-Accompanied Point cloud classification Network (PAPNet)}; built upon $SE(3)$-equivariant convolutions, the PAPNet learns intermediate pose transformations for equivariant features defined on vector fields, which makes the subsequent classification easier (ideally) in the category-level, canonical pose. 
        By adapting existing ModelNet40 and ScanNet datasets to the \emph{single-view, partial} setting, experiment results can verify the necessity of object pose estimation and superiority of our PAPNet to existing classifiers.
		\end{abstract}
		\begin{keyword}
Point Cloud Classification, Rotation Equivariance, Pose Estimation.
		\end{keyword}
	\end{frontmatter}
	
\section{Introduction}
\label{SecIntro}
Semantic analysis of 3D scenes gains increasing popularity with the ubiquitous deployment of depth sensors, where point clouds are usually captured from the sensor fields of view as the most direct representation of 3D shapes. Among various tasks, classification of an input point cloud as one of a set of pre-defined object categories arguably plays the most fundamental role towards semantic understanding of the 3D scenes.

Research on 3D semantic analysis has largely been driven by preparation of benchmarking datasets \cite{Wu20153DSA,Chang2015ShapeNetAI,Song2015SUNRA,Dai2017ScanNetR3,Geiger2012CVPR}. For classification, representative methods such as PointNet \cite{qi2016pointnet}, PointNet++ \cite{qi2017pointnetplusplus}, and DGCNN \cite{Wang2019DynamicGC} follows since the releasing of the ModelNet \cite{Wu20153DSA} and the ShapeNet \cite{Chang2015ShapeNetAI}; for 3D detection (more precisely, 7 degrees-of-freedom object pose estimation), a plenty of methods have been proposed on the KITTI \cite{Geiger2012CVPR}, the SUN-RGBD \cite{Song2015SUNRA}, the ScanNet \cite{Dai2017ScanNetR3}, and other benchmarks. 
Specific settings of these tasks usually follow the original definitions proposed with the benchmarks. Some of these settings could be in controlled, ideal conditions that deviate from those met in practice; consequently, algorithms ranking top on the benchmarks may not work well under practical conditions. For classification, for example, the benchmarking dataset of the ModelNet40 \cite{Wu20153DSA} prepares its training and test instances of object point clouds under the \emph{category-level, canonical poses} \cite{Huang2013FinegrainedSL}.\footnote{Intuitively speaking, one can think that a set of 3D \emph{mug} models are under their category-level, canonical poses when their \emph{handles} are aligned towards one unique direction in the 3D space.} 
Even though such a problem setting is already challenging --- the state-of-the-art methods on the ModelNet40 achieve $\sim 93\%$ accuracies only, researchers find out that addressing the dataset challenges does not always translate as innovations useful for the practical problem of object point cloud classification, since some shortcut learning gives accuracies very close to the state-of-the-art ones, by identifying a sparse set of specially distributed but semantically less relevant points from each object point cloud \cite{geirhos2020shortcut}. 
The degenerate learning is (partially) caused by the less practical setups defined with the benchmarks, since learning setups closer to practical ones are more difficult to be hacked via shortcut learning \cite{geirhos2020shortcut}.

\begin{figure}[t]
\centering
\includegraphics[width=0.980\textwidth]{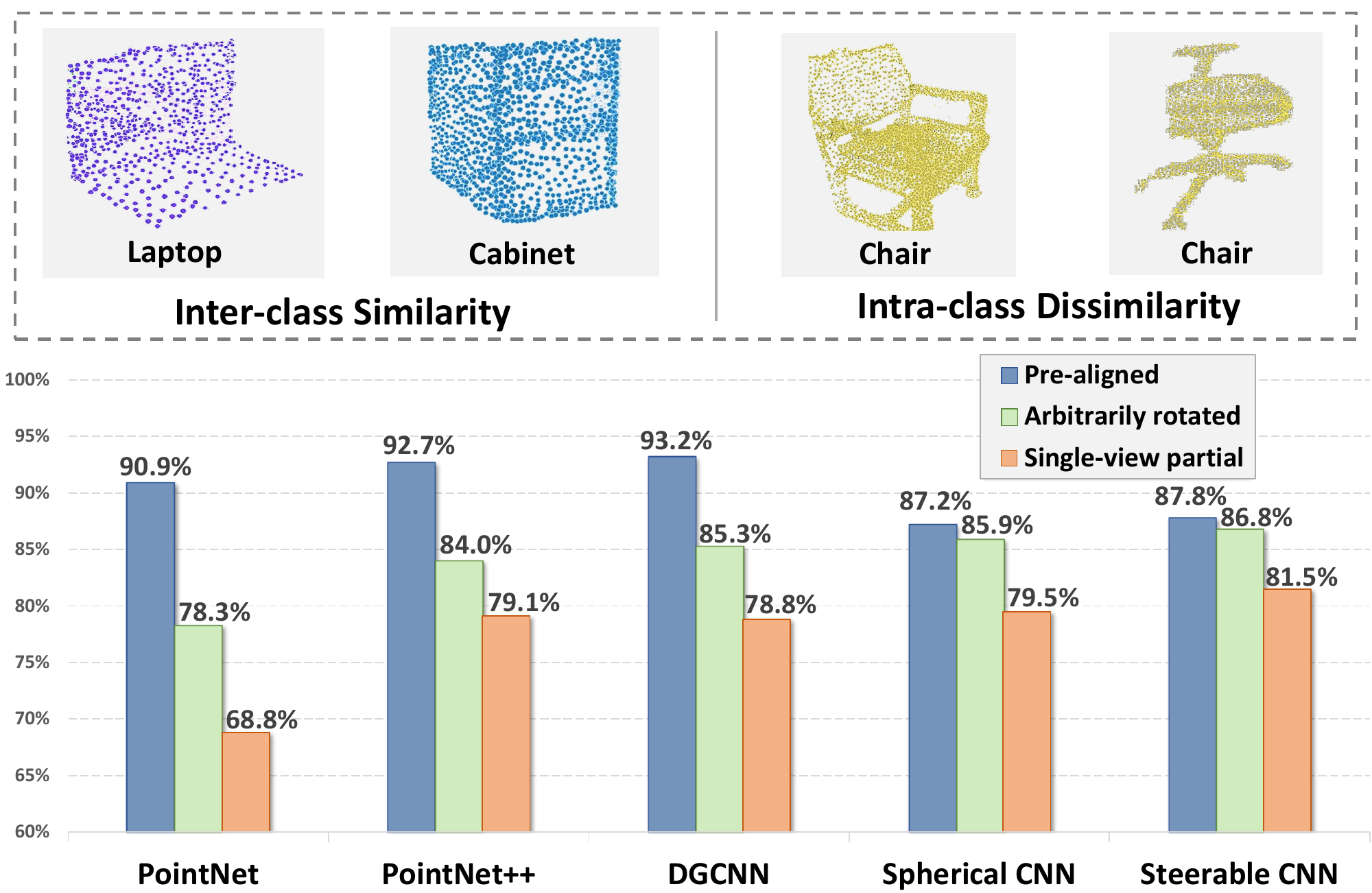}
\caption{Top: four examples from the ScanNet \cite{Dai2017ScanNetR3} to visualize inter-class similarity and intra-class dissimilarity of object point clouds in practice;
Bottom: comparative results of {{five}} popular point cloud classifiers, \ie PointNet \cite{qi2016pointnet}, PointNet++ \cite{qi2017pointnetplusplus}, DGCNN \cite{Wang2019DynamicGC},  Spherical CNN \cite{Esteves2018LearningSE}, and {{3D Steerable CNN \cite{Weiler20183DSC}}}, under the proposed \emph{single-view, partial} setting and existing pre-aligned and arbitrarily posed settings. }\label{fig:intro}
\end{figure}

More recently, the community realizes the less practical setups in existing benchmarks of object point cloud classification, and proposes rotation-augmented versions of these benchmarks by arbitrarily rotating their object instances \cite{chen2019clusternet,Esteves2018LearningSE,zhang-riconv-3dv19}. They also propose strategies coping with the rotation arbitrariness, including implicitly learning object rotations in weakly supervised manners \cite{qi2016pointnet,Yuan2018IterativeTN}, and extracting and learning rotation invariant deep features \cite{chen2019clusternet,zhang-riconv-3dv19,Zhao2019RotationIP}. However, the setups considered in these works are still one step away from the truly practical one, where due to (self-)occlusion, a captured point cloud covers \emph{partial} surface of an object only, instead of the \emph{complete} one considered in \cite{Wu20153DSA, Chang2015ShapeNetAI}.
In other words, the practical object point cloud classification is often under the \emph{single-view, partial} setting, whose examples are shown in the top row of Figure \ref{fig:intro}.  
This subtle difference from complete to partial coverage of the object surface brings significant challenges to point cloud classification, as observed in the bottom plot of Figure \ref{fig:intro}, where the performance of representative methods drops drastically under the single-view, partial setting. Indeed, the semantic category of an object point cloud is defined by its global, topological configuration of local shape primitives; a partial point cloud observed from an arbitrary view captures a subset of local shape primitives, whose classification is less ambiguous only when the distribution of partial point cloud on the whole surface is clearly specified.

We thus argue in this paper that, for effective classification of an arbitrarily posed, partial object point cloud, it is necessary to explicitly learn and predict its pose in the $SE(3)$ space. This technically means that an auxiliary task of \emph{supervised} object pose learning should be accompanied with classification. 
On one hand, classification of object point clouds and pose estimation are relevant tasks, and good performance on one task can benefit the other. 
Specifically, discounting object poses encourages discriminative features for classification, while shape priors of object instances within each class can reduce the difficulty of pose estimation.  
On the other hand, it is not trivial to design a model to jointly estimate object categories and poses on point sets, in view of inherently contradictory characteristics of features desired for both tasks.
In detail, representation learning of object point cloud classification concerns on achieving rotation invariance, whereas rotation equivariant features are demanded for 6D pose estimation.


In this paper, we rely on steerable CNNs \cite{Cohen2017SteerableC,Weiler20183DSC} and propose a new architectural design termed \emph{Pose-Accompanied Point cloud classification Network (PAPNet)}.
The PAPNet starts with a backbone that processes an input point cloud of partial object surface with $SE(3)$-equivariant convolutions, where pose-sensitive, equivariant features defined on the vector fields of each layer are learned with the filters constructed from steerable basis kernels; on top of the backbone, we stack a pose estimation head in parallel with an object classification head;
a key design in our PAPNet is to enforce a transformation of the pose-specific features at the backbone output into the pose-normalized ones in the canonical object pose, where classification can be (ideally) made easier.
Technically, the shared backbone can thus alleviate the dilemma of rotation specific and invariant feature encoding simultaneously; only focus on extracting rotation equivariant features for pose estimation.

In the problem of pose-accompanied classification, class-agnostic pose estimation can be more challenging than classification on single-view partial point clouds; therefore, unreliable pose predictions can lead to a larger feature variation to inhibit object point cloud classification.
To address such a problem, we introduce three simple yet effective strategies to improve feature discrimination -- 1) inspired by the label distribution learning \cite{Geng2013FacialAE, Geng2014HeadPE}, the problem of pose estimation (typically a regression problem, whose labels are intrinsically continuous) is formulated into a supervised classification with a soft target code, tolerant for intra-class feature inconsistency in pose estimation; 2) using ground-truth pose labels with adding random variations in their neighbourhood of pose space to replace pose predictions for arbitrary-to-canonical feature transformation, to reduce feature variations in object classification; and 3) an ensemble of top confidential pose predictions during inference. 
To test the efficacy of our design, existing benchmarks of object point cloud classification are adapted to the single-view, partial setting.
Thorough experiments confirm the necessity of pose estimation for the considered practical setting of object point cloud classification, under which our method greatly outperforms existing ones, especially for the more challenging cross-dataset evaluation.
 
Main contributions are summarized as follows.
 \begin{packed_itemize}
 \item We approach object point cloud classification from a more practical perspective, and propose the single-view, partial setting under which point clouds covering the partial surface of object instances are observed. We discuss the limitations of existing methods, and show that their performance drops drastically under the practical setting.
 \item We propose a baseline method of Pose-Accompanied Point cloud classification Network (PAPNet), 
 which accompanies the classification task with an auxiliary one of supervised object pose learning. Built upon $SE(3)$-equivariant convolutions, a key design in PAPNet is an intermediate transformation of vector-field features to ease the classification by (ideally) making it in the canonical pose space.
 \item To advance the research field, {we adapt existing ModelNet40 and ScanNet benchmarks to the single-view, partial setting.} Thorough experiments show the efficacy of our proposed PAPNet, which performs significantly better than existing methods, especially on the challenging transferability evaluation across datasets.
 \end{packed_itemize}
Datasets and codes will be released after acceptance\footnote{\url{Link-of-datasets-and-codes-to-be-downloaded.}}.

\section{Related Works}
\label{SecRelatedWorks}

In this section, we briefly review existing research that pursues object point cloud classification in practical, real-world conditions, and also the closely related research of object pose estimation.
Discussions on the existing methods of point cloud classification under controlled conditions will be given in Section \ref{SecAlternativeSolution}.

\noindent\textbf{Towards Real-World Object Point Cloud Classification --}
Very few existing works \cite{Uy2019RevisitingPC,Yuan2018IterativeTN} have explored to handle the compound challenges of rotation variations and partially visible shape.
Uy \etal \cite{Uy2019RevisitingPC} propose a realistic point cloud classification benchmark -- the ScanObjectNN, whose setting is similar to ours, but the main differences lie in two folds.
On one hand, object shapes in the ScanObjectNN \cite{Uy2019RevisitingPC} only have arbitrary rotations along the vertical axis, instead of the $SO(3)$ rotation group containing all possible rotation transformations as our setting.
On the other hand, point clouds in the ScanNet \cite{Dai2017ScanNetR3} and the SceneNN \cite{Hua2016SceneNNAS} were obtained by fusing a sequence of depth images with multiple viewpoints, and thus objects segmented by the ScanObjectNN \cite{Uy2019RevisitingPC} can not reflect the single-view partiality of real-world data.
Consequently, although both methods are interested in semantic analysis on partial point clouds, their method concerns more on robust classification performance with noisy point cloud input, while our method treats pose variations as the main challenge.
Yuan \etal \cite{Yuan2018IterativeTN} share a similar observation to handle with partial and unaligned point clouds, but their weakly-supervised learning nature of spatial transformation can hardly guarantee canonical pose transformation
(more details investigated in Sec. \ref{SecAlternativeSolution} and inferior performance to our PAPNet reported in Table \ref{tab:evaluation}).



\vspace{0.1cm}\noindent\textbf{Object Pose Estimation --}
The problem of 6D pose estimation aims to predict object poses (\ie a rotation and translation) in camera space according to a canonical pose.
Existing 6D pose estimation can be categorized into two groups -- instance-level \cite{Xiang2018PoseCNNAC,Li2019DeepIMDI,Wang2019DenseFusion6O,Xu2019WPoseNetDC} and category-level \cite{Wang2019NormalizedOC, Chen2020LearningCS, Tian2020ShapePD}.
In instance-level 6D pose estimation \cite{Xiang2018PoseCNNAC,Li2019DeepIMDI,Wang2019DenseFusion6O,Xu2019WPoseNetDC}, a typical assumption is adopted that CAD models for object instances to be estimated are available.
In this sense, these methods concern more on learning to match partial observation to the CAD model.
In category-level 6D pose estimation \cite{Wang2019NormalizedOC, Chen2020LearningCS, Tian2020ShapePD}, CAD models of each instance are unavailable during training and testing. Such a problem is made more challenging to cope with shape variations of unseen object instances and thus relies on learning a high-quality category-level mean shape.
The pose estimation task in our method falls into the latter group but without estimating the translation and size of object instances compared to existing category-level pose estimation methods, as negative effects of translation and size variations can be eliminated by normalizing the input point cloud to a unit ball in the classification task.
Existing works on category-level 6D pose employ the MLP-based feature encoding, which is less effective to retain rotation information. Our method takes advantage of the equivariance property of steerable convolution to estimate the rotaiton which perform much better.

\section{The Practical Problem Formulation}
\label{SecProbDefinition}

{As examples illustrated in the top row of Figure \ref{fig:intro},} our considered practical setting assumes access to a point cloud $\mathcal{P} = \{ \bm{p}_{i} \in \mathbb{R}^3 \}_{i=1}^N \in \mathcal{X} $ covering \emph{partial} surface of an object. Given a training set $\{ \mathcal{P}_i, y_i \}_{i=1}^M$ of $M$ instances, the task is to learn a classification model $\Phi: \mathcal{X} \rightarrow \mathcal{Y}$ that classifies any test $\mathcal{P}$ into one of $K = |\mathcal{Y}|$ object categories. We define $\mathcal{Y} = \{1, \dots, K\}$ and $y_i \in \mathcal{Y}$ for any $i^{th}$ training instance. 
As discussed in Sec. \ref{SecIntro}, to achieve effective point cloud classification under the single-view, partial setting, it is better to impose an auxiliary supervision for object pose estimation. This amounts to augmenting the training set as $\{ \mathcal{P}_i, y_i, \bm{T}_i\}_{i=1}^M$, where the ground-truth pose $\bm{T}_i \in SE(3)$, $\bm{T} = [\bm{R}|\bm{t}]$ with rotation $\bm{R} \in SO(3)$ and translation $\bm{t} \in \mathbb{R}^3$, applies to $\mathcal{P}_i$ in a point-wise manner and {transforms $\mathcal{P}_i$ into its canonical pose as $\bm{T}_i\mathcal{P}_i$}. Note that the canonical pose of each $\mathcal{P}$ is pre-defined at the category level, which can be obtained in a semi-automatic manner during preparation of training instances \cite{Huang2013FinegrainedSL}. 
In view of difficulty in annotating object poses given visual observation, real world labelled data for 6D pose estimation is typically sparse, which can be alleviated by leveraging synthetic data generated from CAD models with an off-the-shelf renderer such as {{Blender \cite{blender}}}.


In contrast, existing methods either assume an \emph{arbitrarily rotated}, \emph{complete} point cloud $\widetilde{\mathcal{P}} = \{ \bm{p}_{i} \in \mathbb{R}^3 \}_{i=1}^{\widetilde{N}} $ covering the whole surface of an object \cite{chen2019clusternet,zhang-riconv-3dv19,Esteves2018LearningSE}, or assume $\bm{T}\widetilde{\mathcal{P}}$ where $\bm{T}$ 
rigidly transforms $\widetilde{\mathcal{P}}$ into the category-level, canonical pose \cite{qi2016pointnet,qi2017pointnetplusplus,Wang2019DynamicGC}. They correspondingly learn classification models using training sets of either $\{ \widetilde{\mathcal{P}}_i, y_i \}_{i=1}^M$ or $\{ \bm{T}_i\widetilde{\mathcal{P}}_i, y_i \}_{i=1}^M$. Note that in the latter case, $\{ \bm{T}_i \}_{i=1}^M$ are implicitly assumed and are not used for classification learning; in other words, all the training and test instances have been pre-aligned into their canonical poses.

\section{Existing Rotation-Agnostic Methods}
\label{SecAlternativeSolution}

In this section, we discuss existing strategies for classification of \emph{arbitrarily posed, complete} point clouds. Denote $\Phi_{fea}: \mathbb{R}^{\widetilde{N}\times 3} \rightarrow \mathbb{R}^d$ as the feature encoding module, which produces $d$-dimensional feature embedding $\Phi_{fea}(\widetilde{\mathcal{P}})$ for any input $\widetilde{\mathcal{P}}$, and $\Phi_{cls}: \mathbb{R}^d \rightarrow [0, 1]^K$ as the final classifier typically constructed by fully-connected layers.
These methods implement point cloud classification as a cascaded function $\Phi = \Phi_{cls} \circ \Phi_{fea}$.


\noindent\textbf{Weakly Supervised Learning of Spatial Transformation --}
Given an arbitrarily posed input $\widetilde{\mathcal{P}}$, a module $\Phi_{trans}$ parallel to $\Phi_{fea}$ is considered in \cite{qi2016pointnet,Wang2019DynamicGC,Yuan2018IterativeTN} to predict a transformation $\widehat{\bm{T}} = [\widehat{\bm{R}}|\hat{\bm{t}}]$, which is then applied to $\widetilde{\mathcal{P}}$ to reduce the variation caused by pose arbitrariness of $\widetilde{\mathcal{P}}$. 
Learning of $\Phi_{trans}$ is conducted \emph{in a weakly supervised manner}: classification supervision imposed on the network output of classifier $\Phi_{cls}$ propagates error signals back onto $\Phi_{trans}$, whose updating is expected to improve the classification at the network output $\Phi_{cls}$. Such a $\Phi_{trans}$, termed T-Net, is proposed in \cite{qi2016pointnet}, which is further improved as an iterative version in \cite{Yuan2018IterativeTN}
; mathematically, let $\Delta\widehat{\bm{T}}$ denote the predicted transformation update per iteration, and we have the transformation predicted at iteration $t$ as a composition $\widehat{\bm{T}}_t = \prod_{i=1}^t \Delta\widehat{\bm{T}}_i$.
Ideally, the predicted $\widehat{\bm{T}}$ after a certain number of iterations are expected to transform $\widetilde{\mathcal{P}}$ into its canonical pose, such that classification can be made easier on $\widehat{\bm{T}}\widetilde{\mathcal{P}}$; 
however, weakly supervised learning does not guarantee to achieve this, and in practice, the predicted $\widehat{\bm{T}}$ is usually less relevant to canonical pose transformation.

\noindent\textbf{Rotation Invariant Feature Extraction and Learning --}
Rotation invariant feature extraction requires that $[\Phi_{fea}\circ \Phi_{geo}](\bm{R}\widetilde{\mathcal{P}}) = [\Phi_{fea}\circ \Phi_{geo}](\widetilde{\mathcal{P}})$ for any $\bm{R} \in SO(3)$, where a module $\Phi_{geo}$ generates rotation invariant geometric quantities. It can be easily verified that simple quantities such as point-wise norm and angle between a pair of points are rotation invariant, \ie $\|\bm{R}\bm{p}\|_2 = \|\bm{p}\|_2 \ \forall \ \bm{p} \in \mathbb{R}^3$   and $\langle\bm{R}\bm{p}, \bm{R}\bm{p}'\rangle = \langle \bm{p}, \bm{p}'\rangle \ \forall \ \bm{p}, \bm{p}' \in \mathbb{R}^3$.
Higher-order geometric quantities invariant to rotation can be obtained by constructing local neighborhoods around each $\bm{p} \in \mathcal{P}$, which altogether provide rotation invariant input features for subsequent learning via graph networks \cite{qi2017pointnetplusplus,Wang2019DynamicGC}.
Rotation invariant feature learning methods \cite{chen2019clusternet,zhang-riconv-3dv19,You2020PointwiseRN,Zhao2019RotationIP} thus implement $\Phi_{fea}$ by learning point-wise features from these invariant outputs of $\Phi_{geo}$, followed by pooling in a hierarchy of local neighborhoods constructed by graph networks \cite{qi2017pointnetplusplus,Wang2019DynamicGC}.
These methods may also be augmented with global pre-alignment of $\widetilde{\mathcal{P}}$ (ideally) to the canonical pose, by computing the main axes of geometric variations of $\widetilde{\mathcal{P}}$ via singular value decomposition \cite{Zhao2019RotationIP}. We note that these invariant quantities only partially capture the geometric information contained in any $\widetilde{\mathcal{P}}$; consequently, deep features learned from them would not be optimal for classification of $\widetilde{\mathcal{P}}$.

\noindent\textbf{Achieving Invariance via Learning Rotation Equivariant Deep Features --}
Rotation invariance can also be achieved by first encoding the input into rotation equivariant deep features and then converting the rotation equivariant features into the rotation invariant ones.
Typical methods are spherical CNNs \cite{Cohen2018SphericalC,Esteves2018LearningSE}.
Denote a rotation equivariant layer as $\Psi: SO(3) \times \mathbb{R}^{d_{in}} \rightarrow SO(3) \times \mathbb{R}^{d_{out}}$; it processes an input signal $\bm{f}: SO(3) \rightarrow \mathbb{R}^{d_{in}}$ with $d_{out}$ layer filters, each of which is defined as $\bm{\psi}: SO(3) \rightarrow \mathbb{R}^{d_{in}}$.\footnote{Note that both $\bm{f}$ and $\bm{\psi}$, and the filtered responses are defined on the domain of rotation group $SO(3)$; when the layer is the network input, $\bm{f}$ and $\bm{\psi}$ are defined on the domain of a sphere $S^2$.}
Rotation equivariant $\Psi$ has the property $[\bm{\psi} \star [\bm{\mathcal{T}}_{\bm{R}} \bm{f}]](\bm{Q}) = [\bm{\mathcal{T}}_{\bm{R}}' [\bm{\psi} \star \bm{f}]](\bm{Q})$, where $\bm{Q} \in SO(3)$ and $\bm{\mathcal{T}}_{\bm{R}}$ (or $\bm{\mathcal{T}}_{\bm{R}}'$) denotes a rotation operator that rotates the feature function $\bm{f}$ as $[\bm{\mathcal{T}}_{\bm{R}}\bm{f}](\bm{Q}) = \bm{f}(\bm{R}^{-1}\bm{Q})$, and $\star$ denotes convolution on the rotation group; spherical convolution defined in \cite{Cohen2018SphericalC,Esteves2018LearningSE} guarantees achievement of the above property.
A rotation invariant $\Phi_{fea}$ can thus be constructed by cascading multiple layers of $\Psi$, followed by pooling the obtained features over the domain of $SO(3)$.
To implement $\Phi_{fea}(\widetilde{\mathcal{P}})$, one may first cast each point $\bm{p} \in \widetilde{\mathcal{P}}$ onto the unit sphere, with the accompanying point-wise geometry features (\eg the length $\|\bm{p}\|_2$), and then quantize the features to the closest grid of discrete sampling on the sphere.
{To implement $\Phi_{fea}(\widetilde{\mathcal{P}})$, one may need to convert $\widetilde{\mathcal{P}}$ into spherical signals.}
Due to numerical approximations and the use of nonlinearities between layers, the encoder based on spherical CNNs is not perfectly rotation equivariant, which affects the rotation invariance of features obtained by the subsequent pooling \cite{spezialetti2019learning}.
Alternative manners \cite{Thomas2018TensorFN,Weiler20183DSC} exist that directly learn rotation equivariant point-wise filters based on steerable basis kernels in the Euclidean domain.
Similarly, these filters of each layer are stacked together to construct a $SO(3)$-equivariant feature encoding module $\Phi_{fea}$ to generate scalar-, vector-, and tensor- field features, which are normalized to achieve rotation invariance via pooling before fed into the classification module $\Phi_{cls}$.

\vspace{0.1cm}\noindent\textbf{Limitations with Single-View Partial Setting --}
The aforementioned existing methods are originally proposed for classifying point clouds sampled from complete object surfaces; nevertheless, one might be tempted to apply them to the single-view, partial setting considered in the present paper. However, we will analyze that the simple change of settings from complete to partial surfaces brings significant challenges to classification, and may cause either failure or severe performance degradation of the above methods. Indeed, as discussed in Sec. \ref{SecIntro}, semantics of object point clouds are defined by global configurations of local shape primitives; a partial object surface observed from an arbitrary view captures a subset of local shape primitives only, which makes it difficult to specify some semantics that can be clearly defined only when the global configurations are available. 
More specifically, for methods \cite{qi2016pointnet,Wang2019DynamicGC,Yuan2018IterativeTN}, the learned $\widehat{\bm{T}}$ via training $\Phi_{trans}$ in a weakly supervised manner becomes even less relevant to canonical pose transformation when the input is a point cloud $\mathcal{P}$ of partial surface; for both the methods using rotation invariant feature extraction \cite{chen2019clusternet,zhang-riconv-3dv19,You2020PointwiseRN,Zhao2019RotationIP} and those achieving invariance by pooling over learned rotation equivariant deep features \cite{Cohen2018SphericalC,Esteves2018LearningSE, Thomas2018TensorFN, Weiler20183DSC}, a partial $\mathcal{P}$ observed from an arbitrary view is ambiguous for specifying how such a $\mathcal{P}$ is configured globally on the whole surface, and consequently classifiers would be confused between the global and local levels. In addition, we note that pre-alignment of a partial $\mathcal{P}$ via computation of main geometric axes easily causes misalignments.
Figure \ref{fig:intro} verifies the above analysis empirically. More comprehensive experiments are presented in Sec. \ref{SecExps}. 
In view of the limitations of existing methods for the practical challenge of object point cloud classification under the single-view, partial setting, we argue that an explicit, \emph{supervised learning} of object pose from the partial point cloud is necessary. 


\begin{figure*}[t]
\centering \includegraphics[width=1.0\linewidth]{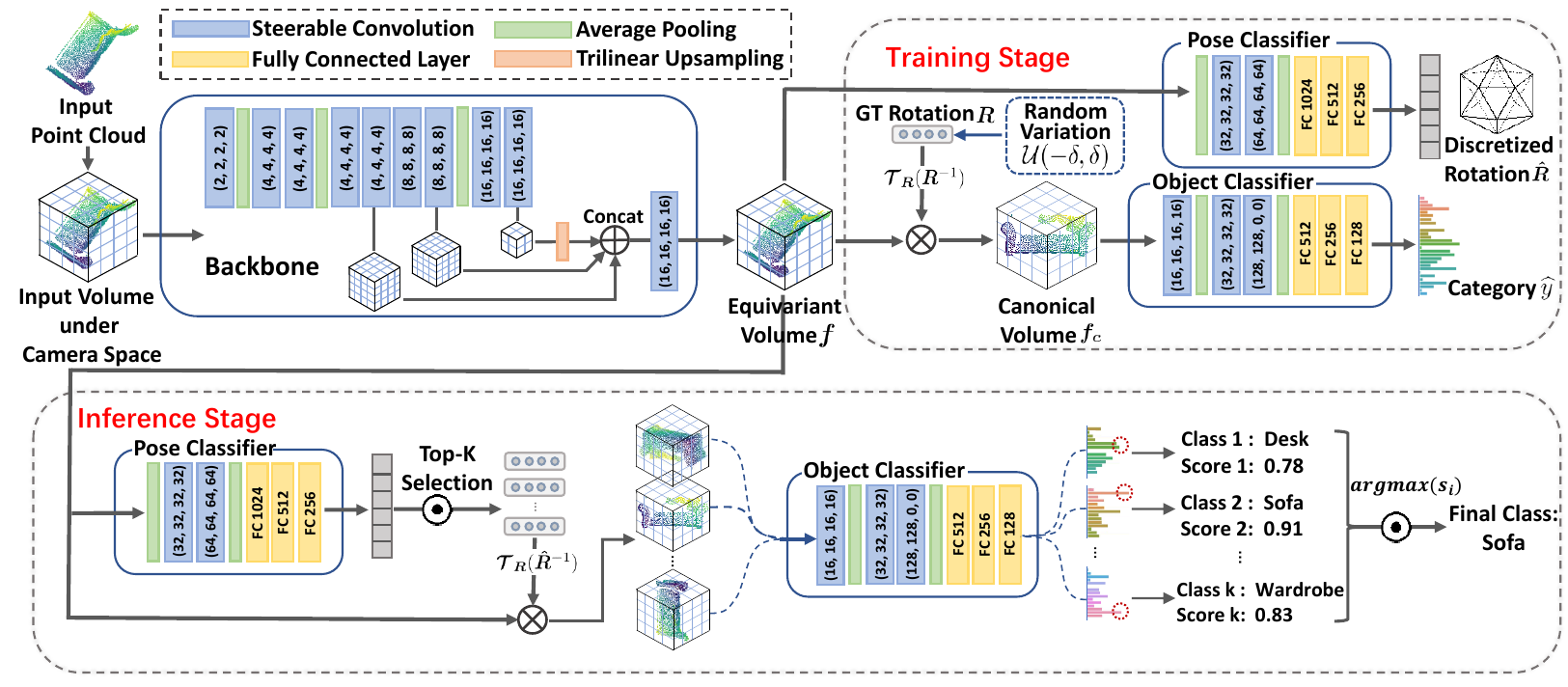}
\caption{Pipeline of our PAPNet on classifying partial point sets with an auxiliary prediction of object pose. The channels of steerable convolution and residual block can be represented by a tuple of numbers, where the position denotes the order of the irreducible features and the corresponding value is the number of features of that order.
}
\label{fig:our_method}
\end{figure*}

\section{Pose-Accompanied Point Classification Net}
\label{SecOurSolutions}


As aforementioned, the global configurations of local shape primitives are desired to specify semantic patterns when only single-view partial point clouds available, which encourages an auxiliary task of supervised regression on object poses to accompany object point cloud classification.
As a result, we propose a novel Pose-Accompanied Point cloud classification Network (PAPNet), whose pipeline is shown in Figure \ref{fig:our_method}, to introduce a key intermediate feature transformation to tackle the dilemma of rotation sensitive and invariant feature encoding for the shared backbone. 

\subsection{$SE(3)$-equivariant Convolution Based on Steerable Basis Kernels}
Our PAPNet relies on the 3D steerable convolution proposed in \cite{Weiler20183DSC}, which guarantees that rigid transformations of objects in the Euclidean space can lead to an equivalent transformation of features in feature space.
To this end, features $\bm{f}(\bm{v}) \in \mathbb{R}^{d_l}$ on position $\bm{v} \in \mathbb{R}^3$ of layer $l$ is represented by a set of scalar-valued and vector-formed {features}, 
where function $\bm{f}:\mathbb{R}^3 \rightarrow \mathbb{R}^{d_l}$ defines the feature space $\mathcal{F}_l$ as a combination of multiple scalar and vector fields.

Each field transforms independently under rigid body motion $\bm{T}= [\bm{R}|\bm{t}]$ as:
\begin{equation}\label{eq:trans_formula}
    [\bm{\mathcal{T}}(\bm{T}) \bm{f}](\bm{v}) := \bm{\rho}_l(\bm{R}) \bm{f}(\bm{R}^{-1}(\bm{v} - \bm{t})),
\end{equation}
where $\bm{\mathcal{T}}$ denotes the transformation operator and $\bm{\rho}_l$ is a representation of $SO(3)$  which describes the rotation behavior of fields in layer $l$.
Since the translation of the observed point cloud can be approximated well by its coordinate mean in the classification task, we translate the observed point cloud to the origin by its coordinate mean and simplify the transformation law in feature space with only rotation $\bm{R}$ as follows:
\begin{equation}\label{eq:trans_formula2}
    [\bm{\mathcal{T}}_{\bm{R}}(\bm{R}) \bm{f}](\bm{v}) := \bm{\rho}_l(\bm{R}) \bm{f}(\bm{R}^{-1}\bm{v}),
\end{equation}
where $\bm{\mathcal{T}}_{\bm{R}}$ denotes the rotation operator.

As any representation of $SO(3)$ can be constructed from irreducible representation of dimension $2k+1$, for $k=0, 1, 2, \ldots, \infty$, which is known as the Wigner-D matrix $D^k(\bm{R})$ of order $k$ \cite{Weiler20183DSC},
the $SO(3)$ representation in layer $l$ can thus be written as:
\begin{equation}
    \bm{\rho}_l(\bm{R}) = \bm{Q}^{-1} \left[ \bigoplus_{i=1}^{F_l} \bm{D}^{k_i}(\bm{R}) \right] \bm{Q},
\end{equation}
where $\bigoplus$ denote the construction of a block-diagonal matrix with blocks $\bm{D}^{k_i}(\bm{R})$, and $\bm{Q}$ is a change of basis matrix. As a result, the $SO(3)$-equivariant features in layer $l$ is a stack of $F_l$ features $f^i(v) \in \mathbb{R}^{2 k_i + 1}$, so that $d_l = \sum_{i=1}^{F_l} (2 k_i + 1)$.

In \cite{Weiler20183DSC}, Weiler \etal derived analytically that the equivariant convolutional kernel between adjacent feature spaces must satisfy the kernel constraint:
\begin{equation}
    \label{eq:kernel_constraint}
    \kappa(\bm{R}\bm{v}) = \bm{\rho}_{l+1}(\bm{R}) \kappa(\bm{v}) \bm{\rho}_l(\bm{R})^{-1},
\end{equation}
where $\kappa : \mathbb{R}^3 \rightarrow \mathbb{R}^{d_{l+1} \times d_l}$ is the convolutional kernel. The solution space formed by this linear constraint can be spanned by the steerable basis kernels \cite{Weiler20183DSC}. Consequently, equivariant convolutional kernel built up by linearly combining these steerable basis kernels using learnable weights guarantee that features in layer $l+1$ transform according to $\bm{\rho}_{l+1}(\bm{R})$ if features in layer $l$ transform according to $\bm{\rho}_l(\bm{R})$. 
Such a rotation equivariant property allows us to use object poses to align features from the arbitrarily posed input point clouds to canonical pose space.

\subsection{Network Architecture}

Dependent on 3D steerable convolutions \cite{Weiler20183DSC}, the proposed PAPNet consists of three modules: a steerable convolutional backbone, a pose classifier, and an object classifier.
Note that, as negative effects of translation $\bm{t}$ in pose $\bm{T}$ can readily be eliminated by normalizing the input point cloud to a unit ball in the classification task, we only concern on estimating rotation $\bm{R}$ of partial point clouds.
With the voxelized representation of a point cloud $\mathcal{P}$ from partial object surface as input, the backbone network processes to generate the rotation equivariant feature $\bm{f}$ which consists of scalar fields and vector fields (see the top left module of Figure \ref{fig:our_method}). 
The output features of the backbone are fed into the pose classifier to output predictions $\widehat{\bm{R}}$,
while the ground-truth $\bm{R}$ with random rotation augmentation $\mathcal U(-\delta, \delta)$ are employed to enforce explicit alignment of the vector-field feature $\bm{f}$ of the backbone into the category-level canonical pose. 
As a result, negative effects of rotation arbitrariness and inter-class shape ambiguity on point cloud classification can be mitigated in the pose-normalized vector-field feature space (see the top right module of Figure \ref{fig:our_method}). 
During inference, given an unseen instance as input, the backbone first generates rotation equivariant feature $\bm{f}$, which is then fed into the pose classifier to output pose probability.
The top $k$ confidential pose candidates $\widehat{\bm{R}}$ are selected for feature transformation on rotation equivariant feature $\bm{f}$, and all the $k$ transformed features are utilized in object classifier to predict semantic category in an ensemble manner (see the bottom row of Figure \ref{fig:our_method}).

\noindent\textbf{Backbone --} {{The backbone module contains ten steerable convolutional layers
and features from three different layers (visualized in Figure \ref{fig:our_method}) are concatenated to enhance the feature discrimination.}}
Such a shared backbone encodes the input into an equivariant feature volume $\bm{f}:\mathbb{R}^3 \rightarrow \mathbb{R}^d$ where the feature vector $\bm{f}(\bm{v}) \in \mathbb{R}^d$ anchored on each voxel $\bm{v} \in \mathbb{R}^3$ is made up of multiple scalars and high-dimensional vectors.
{Direction of the vector features change equivalently with pose varying of input point clouds so as to better retain pose-sensitive information.}

\noindent\textbf{Rotation Classification on Rotation Equivariant Features --}
The rotation estimation branch comprises two steerable convolution layers followed by three fully connected (FC) layers. 
The equivariant feature volume $\bm{f}$ generated by the backbone is fed into the pose classifier to produce a rotation prediction $\bm{\hat{R}}$.
Note that, pose estimation is formulated into a classification problem to assign $\bm{f}$ into one of rotation bins, 
{generated by using a icosahedral group uniformly divide $SO(3)$ space into 60 bin \cite{Yan2012AlmostuniformSO}}, 
whose center is adopted as pose prediction $\bm{\hat{R}}$. 
Such a setting shares the same spirit as $\epsilon$-insensitive Hinge loss in Support Vector Regression \cite{Drucker1996SupportVR} to tolerate unreliable predictions within a pre-defined neighborhood (\ie the width of pose bins) of ground-truth $\bm{R}$. 
To mitigate intra-class feature inconsistency, the typical strategy of using soft label as supervision signal is adopted, which can be generated via {{assigning non-zero probability to multiple nearest neighbours of ground-truth rotation bin}} or adopting label distribution \cite{Geng2013FacialAE, Geng2014HeadPE}. 
It is straightforward that rotation prediction $\bm{\hat{R}}$ can be used to transform the equivariant feature  $\bm{f}$ under the camera space to approach its canonical space by Eqn (\ref{eq:trans_formula2}).
However, we find out that $\bm{\hat{R}}$ can be replaced by ground-truth pose $\bm{R}$ with random variations $\mathcal U(-\delta, \delta)$\footnote{We independently add a noise $\mathcal U(-\delta, \delta)$ to each of the three Euler angles of ground-truth pose $\bm{R}$.} for feature transformation during training, where we obtain the canonical feature volume $\bm{f_c} = \bm{\mathcal{T}}_{\bm{R}}(\bm{R}^{-1}) \bm{f}$ by augmented ground-truth pose $\bm{R}$. Such an replacement can prevent feature inconsistency in object classification from unreliable pose predictions.




\noindent\textbf{Pose-Aligned Object Classification --}
The object classifier consists of three steerable convolution layers and three FC layers.
3D steerable CNN \cite{Weiler20183DSC} achieves rotation invariance by converting all rotation equivariant vector fields into rotation invariant scalar fields before pooling over the Euclidean space. 
Different from \cite{Weiler20183DSC}, our PAPNet transforms the pose-sensitive vector-field features to their canonical pose as input of the classifier and pools the canonical vector fields in the last steerable convolutional layer of the classifier to obtain a {pose-normalized global feature} while preserving discrimination of high-dimensional vector features.
Our motivation lies in, after transformation on equivariant features, feature learning under the category-level canonical space can be made easier owing to less semantic ambiguity.

\noindent\textbf{Loss Functions --}
In our scheme, for each point cloud instance $\mathcal{P}$, we have two types of supervision signals -- the class label $\bm{y}$ and the pose label $\bm{R} \in SO(3)$. 
{For supervising object and pose classifiers, the typical cross entropy loss \cite{qi2016pointnet,Wang2019DynamicGC} is used for both $L_{\text{cls}}$ and $L_{\text{pos}}$,}
and the total loss of our PAPNet is as:
$L_{\text{total}} = L_{\text{cls}} + \lambda L_{\text{pos}}$
where $\lambda$ is a trade-off parameter. 
Moreover, object symmetry is an inevitable problem for pose estimation, as a large number of objects are continuous symmetry or discrete symmetry, \eg bottle, table in the ModelNet40.
We use the method in \cite{Pitteri2019OnOS} to map ambiguous rotation labels to unambiguous ones.

{
\noindent\textbf{Ensemble of Top Confidential Pose Predictions --}
As feature encoding of object classification is made less ambiguous when equivariant feature volume $\bm{f}$ as output of the backbone can be aligned in category-level canonical space,
we adopted an ensemble strategy to improve robustness of the transformed features against unreliable predictions.
To this end, in view of sensitivity of vector-field features to pose variation, we can select top-$k$ confidential pose candidates $\bm{\hat{R_i}}, i=1,...,k$ to transform $\bm{f}$ by
$\bm{f_i} = \bm{\mathcal{T}}_{\bm{R}}(\bm{\hat{R_i}}^{-1}) \bm{f}$; then feed the aligned feature $\bm{f_i}$ into object classifier to obtain $k$ class predictions $\bm{c}_i$ and its corresponding scores $\bm{s}_i$ which response with high activation value at $\bm{c}_i$;  
the final class prediction is generated by $\bm{c}_{out} = \argmax_{\bm{c}_i}(\bm{s}_i)$.


}

\begin{figure}[t]
\centering \includegraphics[width=0.980\linewidth]{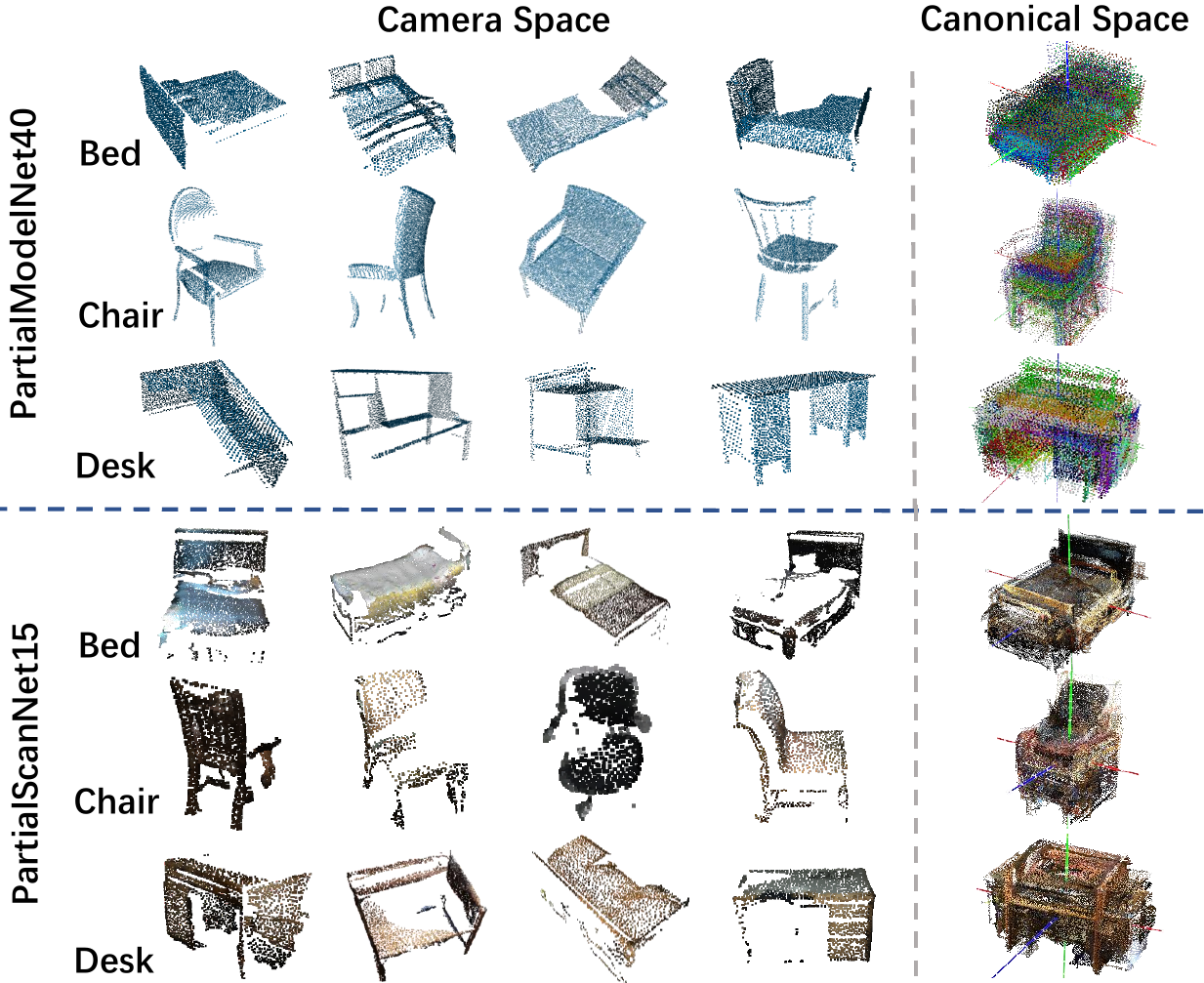}
\caption{Illustration of the PM40 and PS15 datasets adapted from the ModelNet40 \cite{Wu20153DSA} and the ScanNet \cite{Dai2017ScanNetR3}.}
\label{fig:dataset}
\end{figure}

\begin{table}[t]
\caption{Comparative evaluation on classification accuracy (\%) with the PartialModelNet40 (PM40) and the PartialScanNet15 (PS15), as well as transferability evaluation between both datasets.
}
\label{tab:evaluation}
\centering
\resizebox{0.9\columnwidth}{!}{
\begin{tabular}{lccccc}
\toprule
Methods & Input (Size) & PM40 & PS15 & PM40$\rightarrow$PS15 & PS15$\rightarrow$PM40 \\
\midrule
PointNet \cite{qi2016pointnet} & pc ($1024\times3$) & 68.8 & 73.5 & 35.0 & 41.8 \\
DGCNN \cite{Wang2019DynamicGC} & pc ($1024\times3$) & 78.8 & 78.4 & 40.0 & 42.8 \\
PointNet++ \cite{qi2017pointnetplusplus} & pc ($1024\times3$) & 79.1 & 80.4 & 37.2 & 44.1 \\
SimpleView \cite{Goyal2021RevisitingPC}& views ($6\times128^2$) & 79.9 & 84.5 & 47.3 & 49.5 \\
\midrule
TNet-PN++ \cite{qi2016pointnet}& pc ($1024\times3$) & 80.6 & 84.0 & 40.1 & 46.0 \\
ITNet-PN++ \cite{Yuan2018IterativeTN}& pc ($1024\times3$) & 80.9 & 85.2 & 40.1 & 47.5 \\
RRI-PN++ \cite{chen2019clusternet}& pc ($1024\times3$) & 80.8 & 84.1 & 41.7 & 48.8 \\
Spherical CNN \cite{Esteves2018LearningSE}& voxel ($1\times64^2$) & 79.5 & 85.8 & 44.2 & 46.4 \\
Steerable CNN \cite{Weiler20183DSC}& voxel ($1\times64^3$) & 81.5 & 88.1 & 48.5 & 46.5 \\
\midrule
PAPNet (Ours) & voxel ($1\times64^3$) & \textbf{83.5} & \textbf{91.5} & \textbf{56.0} & \textbf{54.6} \\
\bottomrule
\end{tabular}
}
\end{table}

\section{Experiments}\label{SecExps}

\subsection{Single-View Partial Point Cloud Datasets}\label{SecData}
Very few works have explored to generate and release benchmarks of arbitrarily-posed partial point cloud classification, but the available datasets \cite{Wu20153DSA,Uy2019RevisitingPC} lacks object pose annotations, which thus fails to specify the configuration of partial shape on the object surface.
In light of this, the ModelNet40 \cite{Wu20153DSA} and ScanNet \cite{Dai2017ScanNetR3} are adapted to the single-view partial setting with examples shown in Figure \ref{fig:dataset}. 

\noindent\textbf{PartialModelNet40 (PM40) --}
The ModelNet40 \cite{Wu20153DSA} contains 12,311 CAD models belonging to 40 semantic categories which are split into 9,843 for training and 2,468 for testing.
For the PM40, we randomly sampled rotation $\bm{R}$ on $SO(3)$ and translation within $t_x,t_y\sim\mathcal U(-2.0, 2.0), t_z\sim\mathcal U(2.0, 5.0)$, and rendered 10 depth images with corresponding pose labels $[\bm{R}|\bm{t}]$ for each training instance in the ModelNet40 to generate the training set, while its testing set is constructed by randomly sampling one depth image of each testing instance under one arbitrary view. 
These depth images are converted into partial point clouds using the intrinsic parameters of a virtual camera.
In general, the PM40 contains 98,430 training samples and 2,468 testing samples.

\noindent\textbf{PartialScanNet15 (PS15) --}
The ScanNet \cite{Dai2017ScanNetR3} contains 1513 scanned and reconstructed real-world indoor scenes. 
We selected 15 categories with sufficient instances in the scenes and segmented and collected those instances using provided bounding boxes and semantic labels.
With the pose annotations provided by Scan2CAD \cite{Avetisyan2019Scan2CADLC}, the segmented point cloud is first aligned to the canonical space, from which point cloud instances are sampled under ten randomly selected viewpoints.
These point clouds are then transformed into the camera coordinate system.
Since the original ScanNet is generated by fusing a sequence of depth scans and object shapes segmenting from the ScanNet contain redundant surface observed from multiple views,
we employ the hidden point removal (HPR) method \cite{Katz2007DirectVO} to filter out the invisible points due to self-occlusion under a single perspective.
Using HPR may lead to some outliers contaminating the data. We manually filter out samples containing too many outliers, too few observed points, and the mislabeled ones.
In total, the PS15 contains 22,670 training samples and 5,650 test samples belonging to 15 categories.

\subsection{Settings}
\noindent\textbf{Comparative Methods --} We compare representative point classifiers and a number of rotation-agnostic methods mentioned in previous sections with the proposed PAPNet.
Beyond the popular PointNet \cite{qi2016pointnet}, DGCNN \cite{Wang2019DynamicGC} and PointNet++ \cite{qi2017pointnetplusplus}, we take SimpleView \cite{Goyal2021RevisitingPC} as a strong competitor based on multi-view observation generated from point clouds.
The spatial transformation network (TNet) \cite{qi2016pointnet} and iterative transformation network (ITNet) \cite{Yuan2018IterativeTN} are adopted as a weakly-supervised explicit spatial transformation module.
For learning rotation invariant feature representation, we choose the rigorously rotation invariant (RRI) representation proposed in ClusterNet \cite{chen2019clusternet}, which can achieve the state-of-the-art performance with a solid theoretical guarantee.
TNet, ITNet, and RRI are all followed by conventional PointNet++ (PN++) for feature encoding and classification, owing to its best performance among rotation-sensitive point classifiers (see Table \ref{tab:evaluation}).
We employ Spherical CNN \cite{Esteves2018LearningSE} and (3D) steerable CNN \cite{Weiler20183DSC} as competing rotation equivariant methods.
For a fair comparison, generation of point clouds for each object instance is identical in all methods, as well as data augmentation, \ie random $SO(3)$ rotation and jittering.
Default parameters and training strategies suggested in the paper are adopted.

\noindent\textbf{Implementation Details --} We adopt the Adam optimizer with a learning rate of 0.005 for training and train our method for 20 epochs.
{For the soft target code of pose classification, we set a relatively higher value, \ie 0.4, to the ground truth bin, with a value of 0.2 assigned to top 3 nearest bins, based on angular distance as \cite{Huynh2009MetricsF3}.}
Our method contains a total of 1.15M parameters, comparable with other comparative methods (0.12M - 3.59M parameters).

\subsection{Results}
\noindent\textbf{Comparative Evaluation --}
We compare our PAPNet with competing methods on the PM40 and PS15, whose results are shown in Table \ref{tab:evaluation}. 
In general, all rotation-sensitive point classifiers, \ie PointNet, DGCNN and PointNet++, are outperformed by all rotation-agnostic methods under the single-view partial setting, while the SimpleView can perform comparably.
More specifically, both TNet and ITNet can only slightly outperform the baseline PointNet++, indicating that weak supervision on pose transformation can hardly tackle the pose variation in the single-view partial setting.
Spherical CNN and Steerable CNN have better performance than RRI-PN++. 
Such a result can be due to pose-insensitive feature encoding on rotation invariant quantities, which destroys feature discrimination for inter-class similar geometries (see examples in Figure \ref{fig:intro}) and thus less distinguishable, in comparison with implicitly encoding pose variation into features in rotation equivariant methods. 
Our PAPNet gains at least 2.0\% and 3.4\% improvement on classification accuracy over the other methods
on the PM40 and PS15 respectively.
Such an observation demonstrates learning from pose-normalized geometries under the canonical space can mitigate suffering from feature inconsistency with pose variation and partiality.
The performance gap of our PAPNet over the other methods is consistently more significant on the more challenging cross-dataset evaluation
(\eg at least 7.5\% improvement on its baseline Steerable CNN \cite{Weiler20183DSC}), as illustrated in the right two columns of Table \ref{tab:evaluation}, which further supports our claim about the necessity of auxiliary supervised pose estimation on disambiguating intra-class geometric dissimilarity of partial shapes.

\begin{table}[t]
\vspace{-1mm}
{\caption{Ablation studies of our PAPNet on classification accuracy (\%) with the PM40 and the PS15. (Tr. denotes Transformation)}
\label{tab:ablation}}
\centering
\resizebox{0.9\columnwidth}{!}{
\begin{tabular}{lcc||lcc}
\toprule
Methods & PM40 & PS15 & Methods & PM40 & PS15\\
\midrule
PAPNet w/o. pose & 81.6 & 89.8 & Pred Tr. (Reg) & 81.8 & 90.2\\
Cascaded PAPNet & 81.7 & 88.9 & Pred Tr. (Cls) & 81.9 & 90.4\\
Multi-task PAPNet & 81.7 & 89.0 & Pred Tr. (Cls) + Ensem. & 82.0 & 90.8\\
Input Transformation & 82.7 & 91.1 & GT Tr. (Cls) + Ensem. (Ours) & 83.5 & 91.5\\
\bottomrule
\end{tabular}
}
\end{table}
\begin{figure}[t]
\centering 
\includegraphics[width=1.0\linewidth]{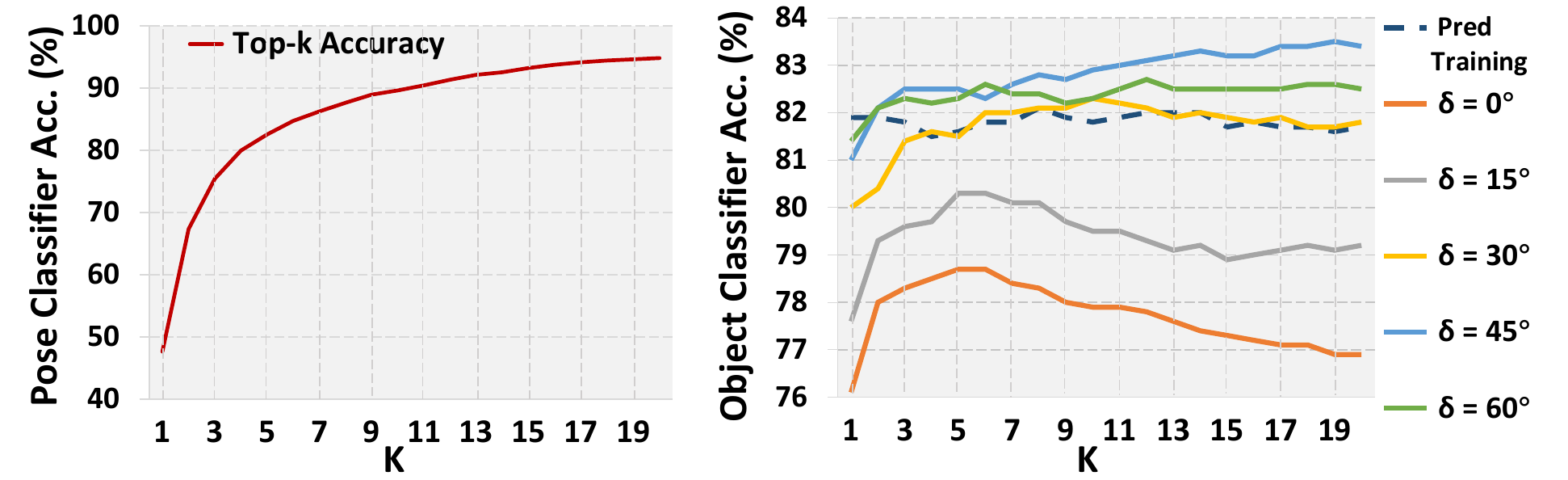}
\caption{Effectiveness of $k$ and $\delta$ in our PAPNet on the PM40.}
\label{fig:ablation}
\end{figure}

\noindent\textbf{Evaluation on Network Structure --}
We compare two degenerated network architectures for joint pose estimation and classification, which are a more fair comparison with the proposed PAPNet, in the perspective of using both pose and class supervision.
The first model is in a cascaded learning manner, where the backbone and pose branch of our PAPNet is used to estimate object pose, while another PAPNet without pose branch (PAPNet w/o. pose) classifies the aligned point cloud using the predicted pose in the first stage.
The second model is organized in typical multi-task learning, sharing a backbone followed by two task-specific headers, which shares the same network architecture as our PAPNet but without feature transformation into canonical space. 
Compared to our PAPNet without pose estimation header, two methods -- cascaded PAPNet and 
multi-task PAPNet can hardly gain any improvement, as shown in Table \ref{tab:ablation}, demonstrating that the problem of joint pose estimation and object classification is not-trivial because of dilemma of contradictory characteristics of features desired in two tasks, which also verify the rationale of network structure of our PAPNet.


\noindent\textbf{Input vs. Feature Transformation --} The only difference between our PAPNet and the ``Input Transformation'' variant in Table \ref{tab:ablation} lies in conducting pose-normalization transformation on the input or the intermediate equivariant features.
The PAPNet with feature transformation performs consistently better on both datasets, as pose-normalized feature encoding for object classification enables the shared backbone to encode complementary information of two tasks with the pose-sensitive vector-field feature.


\noindent\textbf{Pose Regression vs. Classification --} Pose estimation can be formulated into a regression task or a classification task. In Table \ref{tab:ablation}, we see that 
pose classification can achieve a slightly better performance than pose regression, owing to more tolerance (\ie omit small errors within the width of pose bins) to unreliable predictions in pose classification, which encourages us to improve accuracy of pose estimation and alleviate negative effects of unreliable predictions.
  
\begin{figure}[t]
\centering \includegraphics[width=0.70\linewidth]{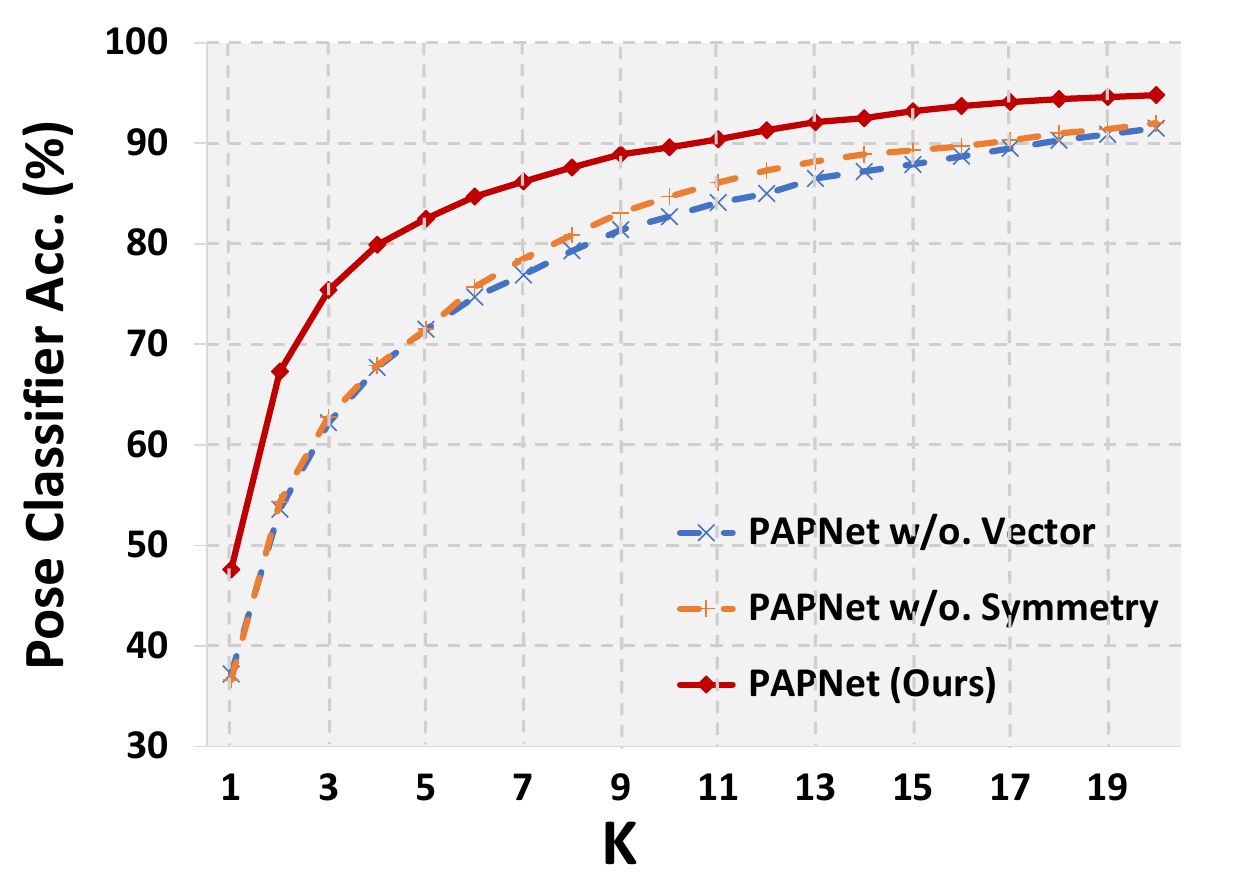}
\caption{Top-$k$ accuracy in pose classification on the PartialModel40. PAPNet w/o. Vector denotes replacing the vector-field features in the last steerable convolution layer of pose classifier with the scalar-field features. PAPNet w/o. Symmetry denotes PAPNet without using the symmetry processing in Sec. \ref{Sec.suppA}.}
\label{fig:pose}
\end{figure}

\noindent\textbf{Pose Prediction vs. Augmented Ground truth for Feature Transformation --} According to Table \ref{tab:ablation}, when predicted pose $\bm{\hat{R}}$ was used for feature alignment during training, there was only a slight improvement even with ensemble strategy (only 0.4\% compared to PAPNet w/o. pose).
Our method using augmented ground-truth pose alignment can gain a 1.9\% improvement by $\mathcal U(-45^{\circ}, 45^{\circ})$, with more results visualized in Figure \ref{fig:ablation}, which further demonstrates that reliable pose can disambiguate feature encoding for object classification.

\noindent\textbf{Effects of Ensemble Strategy --} 
As illustrated in Figure \ref{fig:ablation}, it is observed that, with increasing $k$ in the ensemble strategy, classification accuracy in pose estimation increases, which leads to consistently superior performance on object classification to those without prediction ensembling (\ie $k=0$) regardless of choice of random variation $\delta$ in argumented pose labels. 
Using ground-truth pose for feature alignment in inference, our method obtains 88.2\% and 94.3\% on the PM40 and the PS15 respectively, revealing its upper bound.

\begin{table}[t]
\centering
\caption{Symmetry type of each class in the PartialModelNet40. 'z-inf' denote continuous symmetry along z-axis. 'x/y/z-180' denote 180° symmetry along x-, y- or z-axis.}\label{tab:PM40_sym}
\resizebox{0.70\columnwidth}{!}{
\begin{tabular}{l|l|l|l|l|l} 
\toprule
No. & Class & Symmetry & No. & Class & Symmetry  \\ 
\midrule
0 & airplane & none & 20 & laptop & none  \\
1 & bathtub & z-180 & 21 & mantel & none  \\
2 & bed & none & 22 & monitor & none  \\
3 & bench & none & 23 & night\_stand & z-180  \\
4 & bookshelf & xyz-180 & 24 & person & none  \\
5 & bottle & z-inf & 25 & piano & none  \\
6 & bowl & z-inf & 26 & plant & z-inf  \\
7 & car & none & 27 & radio & xyz-180  \\
8 & chair & none & 28 & range\_hood & none  \\
9 & cone & z-inf & 29 & sink & none  \\
10 & cup & z-inf & 30 & sofa & none  \\
11 & curtain & xyz-180 & 31 & stairs & none  \\
12 & desk & none & 32 & stool & z-inf  \\
13 & door & xyz-180 & 33 & table & z-180  \\
14 & dresser & xyz-180 & 34 & tent & z-180  \\
15 & flower\_pot & z-inf & 35 & toilet & none  \\
16 & glass\_box & xyz-180 & 36 &  tv\_stand & z-180  \\
17 & guitar & y-180 & 37 & vase & z-inf  \\
18 & keyboard & xyz-180 & 38 & wardrobe & xyz-180  \\
19 & lamp & z-inf & 39 & xbox & xyz-180  \\
\bottomrule
\end{tabular}
}
\end{table}

\subsection{Ablation Studies in Pose Classification}
\label{Sec.suppA}

In the ModelNet40, a large number of objects are symmetrical, and therefore one partially observed surface can correspond to multiple or even infinite poses, which can cause ambiguity of mapping from observation to object poses and thus affect accuracy of pose estimation.
The following Map operator proposed in \cite{Pitteri2019OnOS} is adopted to map the pose label of (quasi-)symmetric categories to unambiguous ones as:
\begin{equation}
    \text{Map}(\bm{R}) = \bm{R} \hat{\bm{S}}^{-1}, \: \text{with} \; \hat{\bm{S}} = \underset{\bm{S} \in \mathcal{M}(\mathcal{Y})}{\arg\min} {\| \bm{R} \bm{S}^{-1} - \bm{I}_3 \|_F},
\label{eqn_1}
\end{equation}
where $\mathcal{M}(\mathcal{Y})$ is a set of rigid motion that preserve the appearance or geometry of category $\mathcal{Y}$ and $\bm{I}_3$ is an identity matrix.
For categories with continuous symmetry along the z-axis, such as cup, bowl, and flower pot, $\hat{\bm{S}}$ can be simplified to:
\begin{equation}
  \hat{\bm{S}} = \bm{R}_z(\theta), \: \text{with} \>\> \theta = 
  \text{arctan2}(\bm{R}_{21} - \bm{R}_{12}, \bm{R}_{11} + \bm{R}_{22}) \> ,
\end{equation}
where $\bm{R}_z(\theta)$ denotes rotating objects by angle $\theta$ along the z-axis from the canonical pose. For categories with 180° symmetry along axis $v \in \{x,y,z\}$, $\mathcal{M}(\mathcal{Y}) = \{\bm{I}_3, \bm{R}_v(\pi)\}$ and $\hat{\bm{S}}$ can be easily calculated using Eqn (\ref{eqn_1}).
Symmetry type of each class is shown in Table \ref{tab:PM40_sym}, and corresponding symmetry processing method is adopted in the experiments.
In Figure \ref{fig:pose}, accuracy of pose classifier is greatly improved with symmetry processing (\ie PAPNet vs. PAPNet w/o Symmetry). 
Moreover, experimental results in Figure \ref{fig:pose} can confirm that the equivariance property of vector-field features is beneficial to the pose estimation task (compared to PAPNet w/o. Vector).


\begin{figure}[t]
\centering \includegraphics[width=0.666\linewidth]{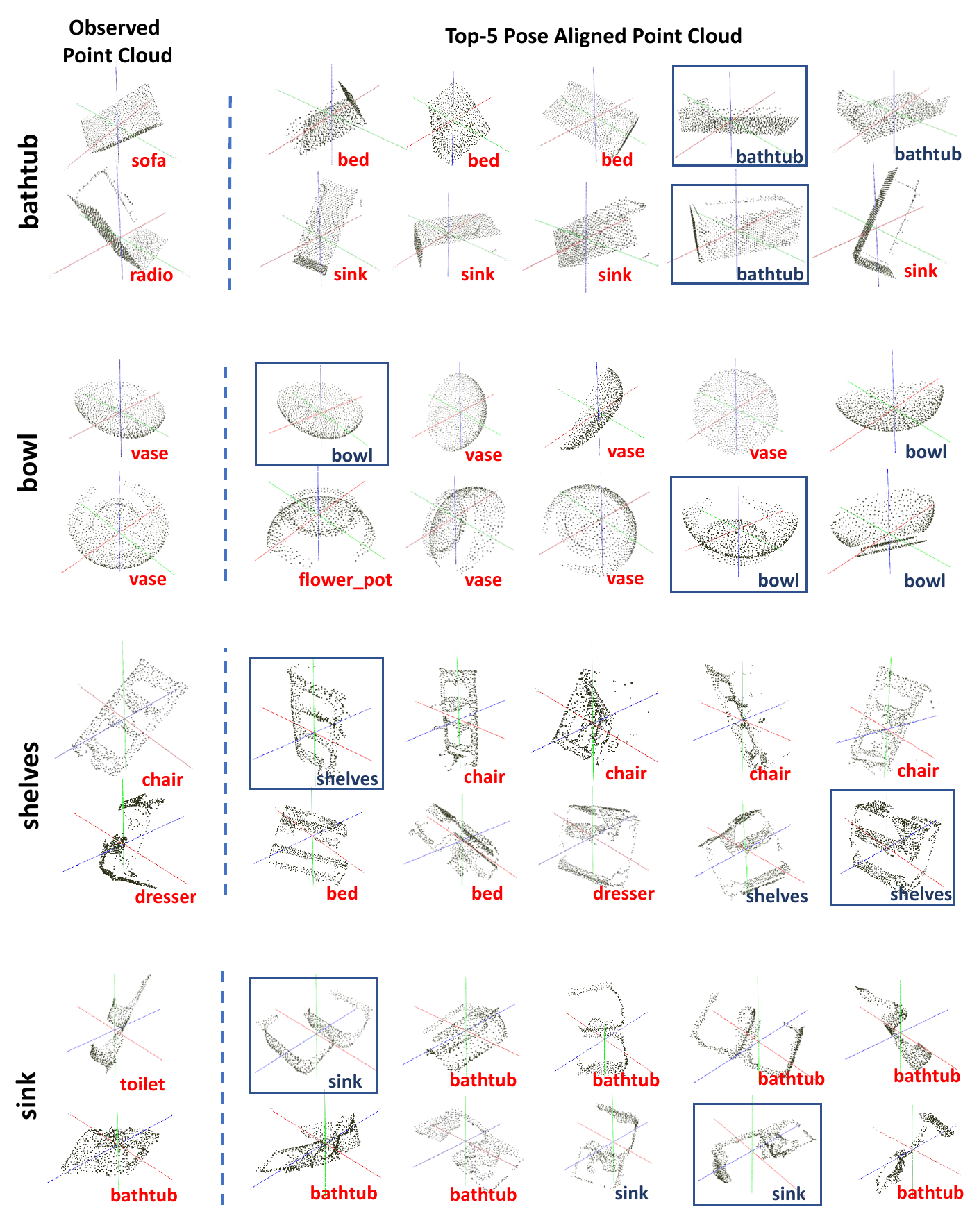}
\caption{Visualization of failure of testing samples of the PAPNet without pose estimation (w/o. Pose) on the left that can be distinguished by our PAPNet on the right. 
The text in red denotes mis-classification of point clouds, while those instances in dark blue text are correctly classified.
The transformed point clouds with top-5 pose predictions can robustly contribute to final classification prediction, which is highlighted in blue bounding boxes.
Note that, pose transformation in our PAPNet is carried out in the feature space, while we apply the pose transformation to the input point cloud for illustrative purpose.
}\label{fig:qualitive}
\end{figure}

\section{Qualitative Results}
Figure \ref{fig:qualitive} visualizes robustness of an ensemble of top-$K$ confidential pose predictions for classifying single-view partial point clouds,  
while partially observed point clouds are easily confused with classes having similar local geometries, \ie without accompanying an auxiliary pose estimation.


\section{Conclusion}

This paper introduces a practical perspective of classification of object point clouds, which desires global configurations to alleviate semantic ambiguity. To this end, a novel classification network accompanied with supervised pose estimation is proposed.
It is found out that accuracy of pose prediction limits classification performance on partial point clouds, which encourages coping with negative effects of unreliable pose predictions.
Experiment results show that feature transformation to category-level canonical space and an ensemble of classification prediction on aligned equivariant features are critically effective, but our method still suffers from inaccurate pose estimation.
{In addition, the contributed scheme might be under adversarial attacks, causing total failure of the whole perception system, which encourages academic researchers and safety engineers to mitigate these risks.
In future, exploration on improving class-agnostic pose estimation is a promising direction.
}


\section*{Acknowledgment}
	This work is supported in part by the National Natural Science Foundation of China (Grant No.: 61771201, 61902131), the Program for Guangdong Introducing Innovative and Enterpreneurial Teams (Grant No.: 2017ZT07X183).


\begin{small}
	\bibliographystyle{elsarticle-num}
	\bibliography{references}
\end{small}

\noindent\textbf{Zelin Xu} received the B.Eng. degree in School of Information Engineering from Guangdong University of Technology, China, in 2019. He is currently pursuing the M.Eng. degree in School of Electronic and Information Engineering from South China University of Technology. His research interests are in computer vision, deep learning, and pattern recognition.

\noindent\textbf{Ke Chen} received the B.Eng. degree in automation and the M.Eng. degree in software engineering from Sun Yat-sen University, China, in 2007 and 2009, respectively, and the Ph.D degree in computer vision from the School of Electronic Engineering and Computer Science, Queen Mary University of London, U.K. He is currently an associate professor in School of Electronic and Information Engineering from South China University of Technology. His current research interests include computer vision, pattern recognition, neural dynamic modeling, and robotic inverse kinematics.

\noindent\textbf{Kangjun Liu} received the B.Eng. degree in College of Mechanical and Vehicle Engineering from Hunan University, China, in 2016. He is currently pursuing the Ph.D. degree in Shien-Ming Wu School of Intelligent Engineering, South China University of Technology. His research interests are in computer vision, deep learning, and pattern recognition.

\noindent\textbf{Changxing Ding} received the Ph.D. degree from the University of Technology Sydney, Australia, in 2016. He is now with the School of Electronic and Information Engineering, South China University of Technology. His research interests include computer vision, deep learning, and medical image analysis.

\noindent\textbf{Yaowei Wang} received the Ph.D. degree in computer science from Graduate University, Chinese Academy of Sciences, in 2005. He is currently an associate researcher at Peng Cheng Laboratory, Shenzhen, China. He was an Assistant Professor with the School of Information and Electronics, Beijing Institute of Technology, and also was a Guest Assistant Professor with the National Engineering Laboratory for Video Technology, Peking University, China. He has been the author or co-author of over 50 refereed journals and conference papers. His research interests include machine learning and multimedia content analysis and understanding. 

\noindent\textbf{Kui Jia} received the B.Eng. degree in marine engineering from Northwestern Polytechnical University, China, in 2001, the M.Eng. degree in electrical and computer engineering from National University of Singapore in 2003, and the Ph.D. degree in computer science from Queen Mary University of London, U.K., in 2007. He is currently a professor in School of Electronic and Information Engineering from South China University of Technology. His research interests are in computer vision, machine learning, and image processing.
	
\end{document}